%% file: main.tex
\documentclass[runningheads]{llncs}
\usepackage{graphicx}
\usepackage{cite}
\usepackage{amsmath,amssymb,amsfonts}
\usepackage{algorithmic}
\usepackage{graphicx}
\usepackage{textcomp}
\usepackage{float}
\usepackage{lipsum}
\usepackage{xcolor}
\usepackage{tabularx,booktabs}
\usepackage{hyperref}
\newcolumntype{C}{>{\centering\arraybackslash}X}
\usepackage{multirow}
\def\BibTeX{{\rm B\kern-.05em{\sc i\kern-.025em b}\kern-.08em
    T\kern-.1667em\lower.7ex\hbox{E}\kern-.125emX}}
\usepackage[font=small,labelfont=bf]{caption} % Required for specifying captions to tables and figures
\usepackage[subrefformat=parens]{subcaption}
\begin{document}
\title{Interpreting Deep Glucose Predictive Models for Diabetic People Using RETAIN\thanks{This  work  is  supported  by  the  "IDI  2017"  project  funded by the IDEX Paris-Saclay, ANR-11-IDEX-0003-02.}}

\titlerunning{Interpreting Deep Glucose Predictive Models Using RETAIN}

\author{Maxime De Bois\inst{1}\orcidID{0000-0002-4181-2422} \and Moun{\^i}m A. El Yacoubi \inst{2}\orcidID{0000-0002-7383-0588} \and Mehdi Ammi\inst{3}\orcidID{0000-0003-1763-4045}}

\authorrunning{M. De Bois et al.}
\institute{CNRS-LIMSI and Universit{\'e} Paris Saclay, Orsay, France \\\email{maxime.debois@limsi.fr} \and Samovar, CNRS, T{\'e}l{\'e}com SudParis, Institut Polytechnique de Paris, \'{E}vry, France \email{mounim.el\_yacoubi@telecom-sudparis.eu}  \and Universit{\'e} Paris 8, Saint-Denis, France \\\email{ammi@ai.univ-paris8.fr}}

\maketitle              % typeset the header of the contribution
\input{0-Abstract}
\input{1-Introduction}

\input{2-RETAIN}
\input{3-Methods}
\input{4-Discussion}
\input{5-Conclusion}

\end{document}

%% file: 0-Abstract.tex
\begin{abstract}
Progress in the biomedical field through the use of deep learning is hindered by the lack of interpretability of the models. In this paper, we study the RETAIN architecture for the forecasting of future glucose values for diabetic people. Thanks to its two-level attention  mechanism, the RETAIN model is interpretable while remaining as efficient as standard neural networks. 

We evaluate the model on a real-world type-2 diabetic population and we compare it to a random forest model and a LSTM-based recurrent neural network. Our results show that the RETAIN model outperforms the former and equals the latter on common accuracy metrics and clinical acceptability metrics, thereby proving its legitimacy in the context of glucose level forecasting. Furthermore, we propose tools to take advantage of the RETAIN interpretable nature. As informative for the patients as for the practitioners, it can enhance the understanding of the predictions made by the model and improve the design of future glucose predictive models.

\end{abstract}

%% file: 1-Introduction.tex
\section{Introduction}

Diabetes is undoubtedly one of the major diseases of the modern world as it has been inputed a total of 1.5 million deaths in 2012 \cite{world2016global}. The every day challenge faced by diabetic people is the regulation of their blood glucose level which is troubled by either the non-production of insulin (type-1 diabetes) or the increasing body resistance to its action (type-2 diabetes). Diabetic people are at risk of facing short terms complications (e.g., coma, death) due to their glycemia falling too low (hypoglycemia) and also long-term complications (e.g., cardiovascular diseases, blindness) when it gets to high (hyperglycemia).

To help the patients coping with their disease, a lot of technological efforts have been made in the recent years. For instance, by enabling the diabetic patient to forgo the use of lancets to get his or her glucose level, continuous glucose monitoring (CGM) devices (e.g., FreeStyle Libre \cite{olafsdottir2017clinical}) are getting more and more common. Besides, we are witnessing the rise of coaching applications specifically made for diabetic people (e.g., mySugr \cite{rose2013evaluating}). From a research perspective, current endeavors are focused towards the building of glucose predictive models. Using past glucose values, carbohydrate (CHO) intakes, insulin infusions, and more, the models forecast the future glucose values at horizons varying from 30 minutes (short-term) to 120 minutes (long-term) \cite{oviedo2017review}.

Thanks to the increasing availability of data and the access to more computing power, the glucose predictive models are shifting from rather simple models (e.g., autoregressive models \cite{sparacino2007glucose}), to more complex algorithms from the machine learning and deep learning field. Daskalaki \textit{et al.} have demonstrated the superiority of feed-forward neural networks over the autoregressive models in the context of short-term glucose forecasting \cite{daskalaki2012real}. Georga \textit{et al.} explored the usability of extreme learning machines for short-term glucose prediction as well \cite{georga2015online}. Recurrent neural networks have recently generated a lot of interest because of their temporal nature, making them particularily suitable for the task of predicting future glucose values \cite{mirshekarian2017using, debois2019prediction}. As time-series can be seen as one-dimension images, convolutional neural networks, which are very popular in the image recognition community, have also been tried out for the forecasting of future glucose values \cite{li2019convolutional}. 

Even though deep models can be effective for the task of glucose prediction, they have a sizable downside: the deeper the model, the more difficult it is to understand its behavior. This is especially an issue for biomedical applications for which it is important to be able to interpret the models in order to understand why a prediction is being made. To address this issue, Georga \textit{et al.} showed that Random Forests (RF), while being highly interpretable, can achieve good performances for the task of glucose prediction \cite{georga2012predictive}. 

Recently, Choi \textit{et al.} proposed a neural network, called RETAIN, specifically designed for healthcare applications dealing with temporal inputs. Featuring a two-level attention mechanism, the model is meant to be as performant as standard neural networks while being interpretable. This property is highly valuable for the prediction of future glucose values. On one hand, it would help the practitioner design better and safer models by providing a better error analysis tool. On the other hand, for the patient, it would help him or her understand his or her desease better.

In this work, we study the use of the RETAIN architecture for the challenging task of the forecast of future glucose values for diabetic people. In particular, we adapt its interpretability feature to regression problems and propose several analysis and visualization tools to interpret the predictions made by the model.

The rest of the paper is structured as follows. First, we describe the RETAIN architecture and how the predictions are interpreted from it. Then, we describe the overall experimental methodology. Finally, we provide  the results and analysis of the experiments before concluding.

% Finally, attention-based neural networks, popular in the natural language processing field \cite{}

% The traditionnal glucose predictive models are quite simple (e.g., autoregressive models \cite{sparacino2007glucose}), there is a current shift towards more complex algorithms, and, in particular \textit{machine-learning} \textit{deep-learning} algorithms. In 2015, De Paula \textit{et al.} proposed the use of Gaussian processes paired with reinforcement learning for the prediction and regulation of bloog glucose level \cite{de2015controlling}. 

% Finally, recurrent neural networks (RNN) have shown a lot of interest in the field \cite{zarkogianni2011insulin}, and in particular those with long short-term memory (LSTM) units \cite{mirshekarian2017using,debois2018study}.

% While displaying better performances \cite{de2019study}, they have also the downside of being less interpretable. This is particularily a concern in the biomedical field as we need to be able to understand how a model behaves and how a prediction is computed. 

% RandomForest \cite{georga2012predictive,li2016smartphone}

%% file: 2-RETAIN.tex
\section{RETAIN}

This section presents the RETAIN architecture that has been previously introduced in \cite{choi2016retain} and its interpretation for time-series forecasting, and in particular for glucose prediction.

\subsection{Architecture}

\begin{figure*}[!ht]
    \centering
    \includegraphics[width=\textwidth]{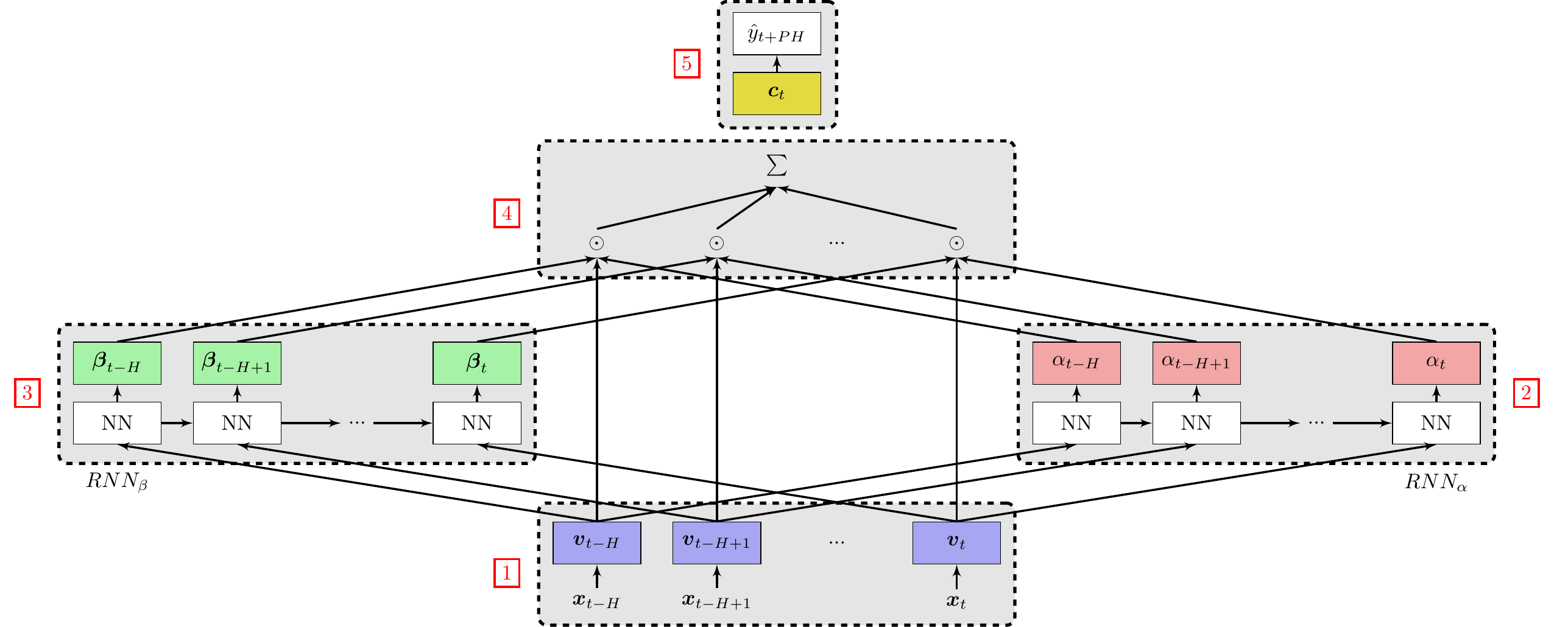}
    \caption{Graphical overview of the RETAIN model. \textbf{Step 1}: The input signals are transformed into embeddings. \textbf{Step 2}: Time-level attention weights are computed from the embeddings. \textbf{Step 3}: Variable-level attention weights are also computed from the embeddings. \textbf{Step 4}: Using the attention weights, the context vector is computed. \textbf{Step 5}: The prediction is made from the context vector.}
    \label{fig:retain}
\end{figure*}

Most of the efficiency of the RETAIN model comes from its two levels of attention: the \textit{time-level} attention (also called \textit{visit-level} attention \cite{choi2016retain}), and the \textit{variable-level} attention. The general attention mechanism comes from the natural language processing field where it enables the model to understand relationships between words in a sentence \cite{vaswani2017attention}. Here, when dealing with temporal inputs (e.g., time-series, events), while the time-level attention makes the network focus on specific time-steps, the variable-level attention enphasizes specific input within the time-steps.

The predictions of the RETAIN model are made following five different steps for which Figure \ref{fig:retain} provides a graphical representation. In the following, $t$ refers to the current time-step the prediction is made, $r$ to the number of different input variables, $H$ to the size of the history (the number of past values for every input variable), $PH$ to the prediction horizon, the subscript $i \in [t-H, t]$ to the \textit{i}-th time-step, and the subscript $j \in [1, r]$ to the $j$-th input variable.
\begin{itemize}
    \item \textbf{Step 1}: Each time-step input vector $ \boldsymbol{x}_i$ is linearly transformed into a learnable embedding $\boldsymbol{v}_i$ following: $\boldsymbol{v}_i = \boldsymbol{W}_{emb} \boldsymbol{x}_i$. 
    \item \textbf{Step 2}: These embeddings are given as inputs to a recurrent neural network, $RNN_\alpha$, which outputs the time-level attention weights $\alpha_i$ (see \cite{choi2016retain} for more details).
    \item \textbf{Step 3}: Similarly, the embeddings are fed into a second recurrent neural network, $RNN_\beta$, which computes the variable-level attention weights $\boldsymbol{\beta}_{i}$ (see \cite{choi2016retain} for more details).
    \item \textbf{Step 4}: Using both attention weights, the context vector $\boldsymbol{c}_t$ is computed following: $\boldsymbol{c}_t = \sum_{i=t-H}^t \alpha_i \boldsymbol{\beta}_i \odot \boldsymbol{v}_i$.
    \item \textbf{Step 5}: The predictions of the model are made by linearly transforming the context vectors: $\hat{y_{t+PH}} = \boldsymbol{W}\boldsymbol{c}_t + b$.
\end{itemize}

The only difference between our architecture and the original one is that we do not compute the attention weights in reverse time-order (Steps 2 and 3) \cite{choi2016retain}, but rather in forward order, as the latter yielded better performances for our application. 

\subsection{Interpretation}

\label{sec:retain_interpret}

In their original paper, the authors of RETAIN propose a way to interpret the outputs of the RETAIN model in the context of multiclass classification. We propose here an adaptation of the methodology to regression problems.

By going through the different operations made in the model, we can express the prediction $\hat{y}_{t+PH}$ in this form:

\begin{equation}
\hat{y}_{t+PH} = \sum_{i=t-H}^{t} \sum_{j=1}^{r} x_{i,j} \alpha_i \boldsymbol{W} (\boldsymbol{\beta}_i \odot \boldsymbol{W}_{emb}[:, j]) + b
\label{eqn:prediction}
\end{equation}

We can then express the contribution $\omega(\hat{y}_{t+PH},x_{i,j})$ of the $x_{i,j}$ input feature on the prediction $\hat{y}_{t+PH}$ as follows:

~
\begin{equation}
\omega(\hat{y}_{t+PH},x_{i,j}) = \alpha_i \boldsymbol{W} (\boldsymbol{\beta}_i \odot \boldsymbol{W}_{emb}[:, j]) ~ x_{i,j}
\label{eqn:contrib}
\end{equation}

While the contributions in this form are useful to analyze an individual sample, they are not very practical if we want to perform further analysis and statistics. Instead, we propose to look at the absolute normalized contribution values $\omega_{AN}(\hat{y}_{t+PH},x_{i,j})$: 

\begin{equation}
% \omega_{AN}(\hat{y}_{t+PH},x_{i,j}) = \frac{\left\lvert \omega(\hat{y}_{t+PH},x_{i,j}) \right\rvert}{\underset{i,j}{\max} (\left\lvert \omega(\hat{y}_{t+PH},x_{i,j}) \right\rvert)}
\omega_{AN}(\hat{y}_{t+PH},x_{i,j}) = \frac{\left\lvert \omega(\hat{y}_{t+PH},x_{i,j}) \right\rvert}{\sum_{i=t-H}^t \sum_{j=1}^r \left\lvert \omega(\hat{y}_{t+PH},x_{i,j}) \right\rvert}
\label{eqn:contrib_norm}
\end{equation}

Taking the absolute values makes the computation of the mean contribution accross the samples more representative of the overall contributions, preventing positive and negative contributions from canceling each other. Normalizing the contributions makes the contributions independent from the prediction value itself, enabling a better comparison between samples.

%% file: 3-Methods.tex
\section{Methods}
\label{sec:methods}

\subsection{Experimental Data}

In this study, we use the IDIAB dataset whose collection has been approved by the french ethical comittee (ID RCB 2018-A00312-53). It is made of data coming from 5 type-2 diabetic patients (4F/1M, age 58.8 $\pm$ 8.28 years old, BMI 30.76 $\pm$ 5.14 $kg/m^2$, HbA1c 6.8 $\pm$ 0.71 \%) that have been monitored for 31.8 $\pm$ 1.17 days in free-living conditions. Whereas their glucose level (in $mg/dL$) was recorded through the use of the FreeStyle Libre continuous glucose monitoring device, data related to CHO intakes (in $g$) and insulin (in units) infusions were manually reported through the mySugr (mySugr GmbH) smartphone coaching app for diabetes.

% Approved by the french ethical comittee (ID RCB 2018-A00312-53), the corpus has been collected from 5 type-2 diabetic patients from which data such as glucose values, insulin infusions, and CHO intakes have been extracted during 8 weeks. The details of the experimental settings can be found in our previous study in \cite{debois2019prediction}.

\subsection{Models}

In this study, we build global glucose predictive models. Whereas personalized glucode predictive models are often more accurate, global models have the advantage of being easier to train by avoiding overfitting thanks to more training data. 

We describe here the preprocessing and training steps of the different models used in this study.

\subsection{Data Preprocessing}

After splitting the patients into four training patients and one testing patient, we have splitted each training patient's data into a training set and a validation set following a 75\%/25\% distribution.

To predict the future glucose values at an horizon of 30 minutes, the models are given as inputs the histories of glucose values, insulin infusions, and CHO intakes of the past 3 hours. For every patient, these inputs are standardized (zero mean and unit variance) w.r.t their respective training set. 

\subsection{Model Training}

The Random Forest (RF) model \cite{svetnik2003random} is one of the two baseline models used in this study. Its main strength is that it provides generally good performances while being easily interpretable. Here, a forest of size 100 is fitted using the mean-squared error (MSE) criterion. The minimum number of samples per leaf has been set to 25 to reduce the overfitting of the model to the training set.

Our second baseline, the LSTM model, has been implemented with an architecture that matches the computational complexity of the RETAIN model described below. In particular, every time-step input variables are embedded into a learnable vector of size 64. These embeddings are then given to a 2-layer LSTM model with 128 units per layer. The latter has been trained to minimize the MSE loss function with the Adam optimizer (learning rate of $10^{-3}$, mini-batch size of 50). To prevent the overfitting of the network to the training set, the early stopping methodology (patience of 25) has been used. 

As for the LSTM model, the RETAIN model has an embedding size of 64. Both RNN\textsubscript{$\alpha$} and RNN\textsubscript{$\beta$} are made of one layer of 128 LSTM units. Similarily, the Adam optimizer (same learning rate and mini-batch size) with the early stopping methodology was used to fit the model.

All the hyperparameters have been optimized by grid search on the validation set on a subspace delimited by manual search.

\subsection{Evaluation}

The models have been evaluated with a 4-fold cross-validation on the training patients followed by a leave-one-(patient)-out cross-validation.

Four different metrics have been used: the Root-Mean-Squared Error (RMSE), the Mean Percentage Absolute Error (MAPE), the Time Lag (TL), and the Continuous Glucose-Error Grid Analysis (CG-EGA).

Both the RMSE and MAPE metrics give a measure of the accuracy of the prediction. The TL metric provides an estimate of the time gained by doing the prediction and is computed as the time-shift (in minutes) that maximizes the correlation between the true and the predicted glucose values. Finally, the CG-EGA measures the clinical acceptability of the predictions \cite{kovatchev2004evaluating}. By analyzing both the prediction accuracy and the accuracy of the variation between two consecutive predictions, the CG-EGA classifies the prediction either as an accurate prediction (AP), a benign error (BE), or an erroneous prediction (EP). For a model to be clinically acceptable, it needs to have high AP and low EP rates.

%% file: 4-Discussion.tex
\section{Results \& Discussion}

\subsection{Experimental Results}

\input{tables/results.tex}
\input{tables/results_cg_ega.tex}

The performances of the three models are shown in Table \ref{table:res} and Table \ref{table:res_cg_ega}. With an average deterioration of 1.4\% in RMSE/MAPE/TL when compared to the LSTM model, the RETAIN model displays a comparable prediction accuracy. Its clinical acceptability is also very similar to the LSTM model.

When compared to the RF model, the RETAIN model shows an improvement of 8.5\%, 8.4\%, 20.7\% in the RMSE, MAPE, and TL metrics respectively. It also has a better clinical acceptability with a lower EP rate (-9.7\%) which comes at the cost of a slightly lower AP rate (-3.3\% of the remaining room for improvement).

Overall, these results are showing that the RETAIN model is a legitimate model for the task of glucose prediction.

\subsection{Interpreting the RETAIN Model}

The real strength of the RETAIN model, however, lies in its interpretability. We propose here several different visualization tools for the analysis of the behavior of the RETAIN model. To ease the reading, we will refer to the contribution as the absolute normalized contribution, presented in Section \ref{sec:retain_interpret}.

\begin{figure}
    \centering
    \includegraphics{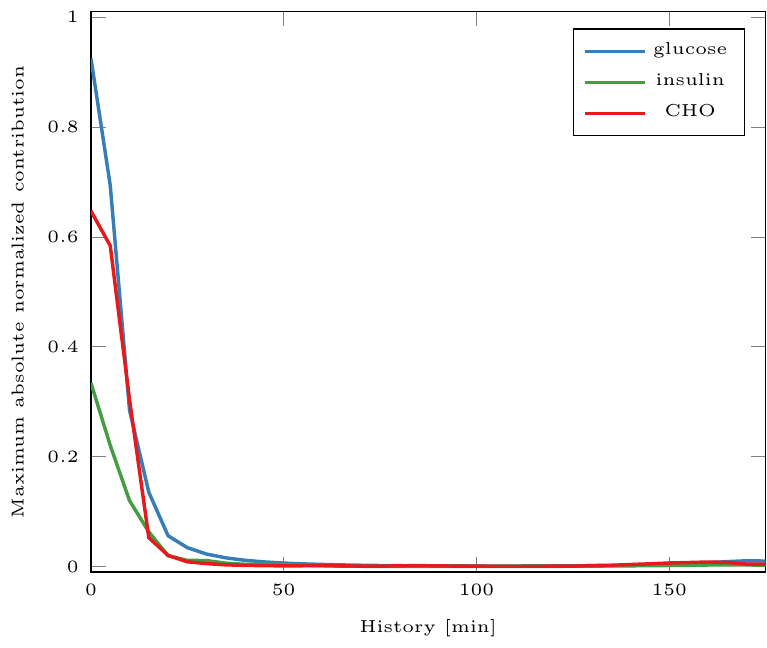}
    \caption{Maximum absolute normalized contribution of the input signals (history of glucose, insulin, and CHO).}
    \label{fig:maxabs_contrib}
\end{figure}

First, by looking at the individual maximum contribution of the input variables, we can see if each of them has ever contributed significantly to the prediction. Figure \ref{fig:maxabs_contrib} plots the maximum contribution of the model inputs related to the 3-hour histories of glucose values, insulin infusions, and CHO intakes. We can see that the the older an input value is, the less contribution it has. The decrease in the contribution is faster for the insulin and CHO signals (close to zero after 30 minutes) than for the glucose signal (close to zero after 60 minutes). This suggests that it is not usefull in this context to use histories that are longer than one hour. Reducing the number of past values inputed to the model should increase the performances by making it harder to overfit and should reduce the training time. Such an analysis is not possible with a standard LSTM model.

\begin{figure*}
    \centering
    \begin{subfigure}[t]{\textwidth}
        \centering
            \includegraphics[width=\linewidth]{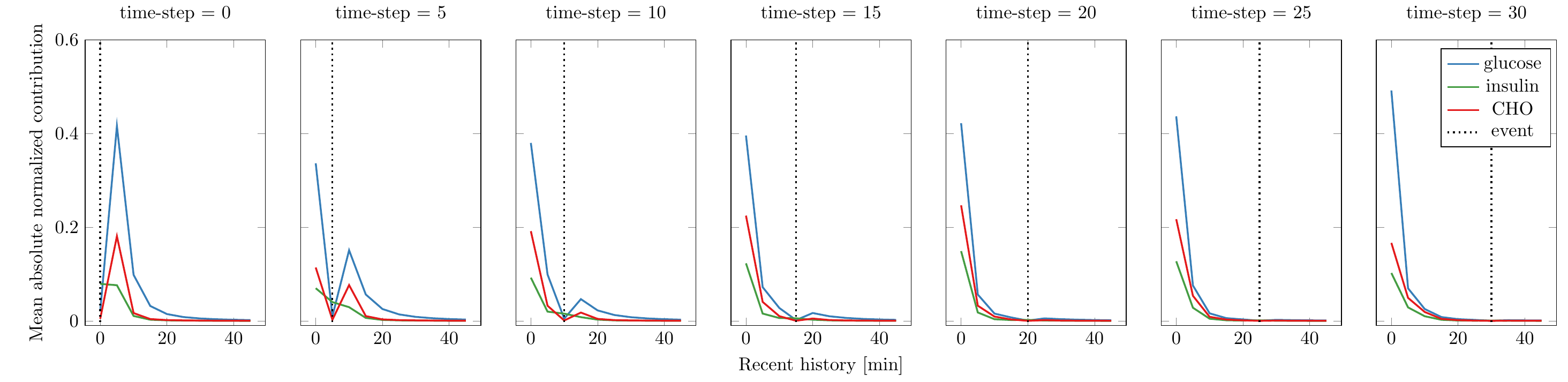}
        \caption{Event: insulin infusion}
        \label{subfig:event_ins}
    \end{subfigure}%

    \begin{subfigure}[t]{\textwidth}
        \centering
        \includegraphics[width=\linewidth]{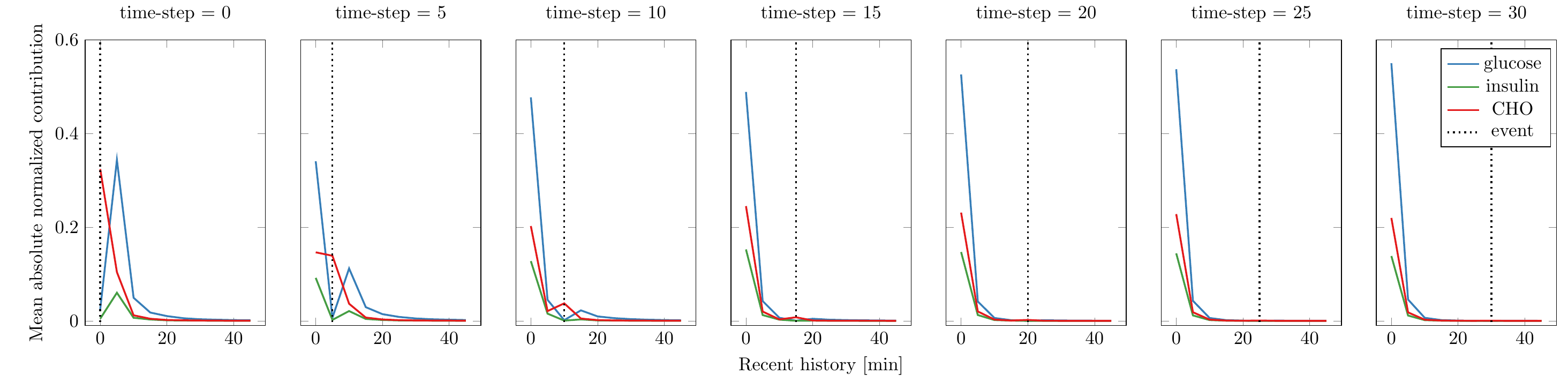}
        \caption{Event: CHO intake}
        \label{subfig:event_cho}
    \end{subfigure}
    \caption{Mean evolution through time of the absolute normalized contribution of the input signals (history of glucose, insulin, and CHO) after the occurence of an event: Figure \ref{subfig:event_ins}, insulin infusion; and Figure \ref{subfig:event_cho}, CHO intake.}
    \label{fig:time_contrib}
\end{figure*}

\begin{figure}
    \centering
    \includegraphics{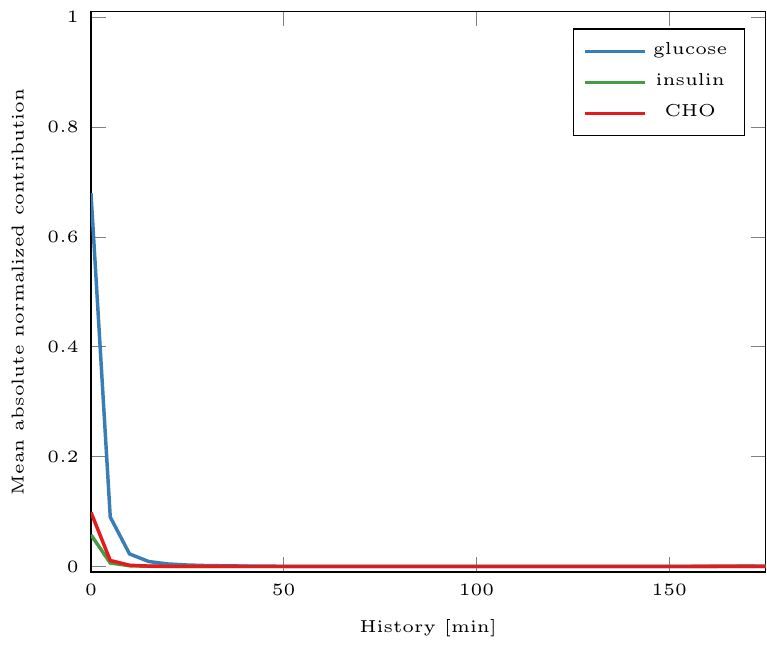}
    \caption{Mean absolute normalized contribution of the input signals when no event (CHO intake or insulin infusion) occured in the last hour.}
    \label{fig:mean_no_event}
\end{figure}

From a different perspective, we can look at the behavior of the model when an event occurs. Figure \ref{fig:time_contrib} depicts the behavior of the model following the occurence of two different events: insulin infusions and CHO intakes. We can compare these plots to the mean contribution when no event has occured in the last hour with Figure \ref{fig:mean_no_event}.

When either one of the events occurs, we can see that the glucose value that has the most importance is not the current glucose value, but the previous one (which is the value 5 minutes before the event). This specific value keeps a relative high importance as the time moves on. This shows that, when an event occurs, the model uses the last glucose value before the event as a value of reference. On the other hand, for both insulin and CHO signals, when their respective event occurs, the contribution of the value of the event is relatively high for the next 20 minutes. However, after this time, the contribution of the event is close to zero and the mean contribution profile becomes similar to the one for which no event has occured in the past hour, depicted by Figure \ref{fig:mean_no_event}.

%% file: tables/results.tex
% (.*)\t(.*)\t0,(\d\d)(.*)\t0,(\d\d)(.*)\t0,(\d\d)(.*)
% \1 & \2 & \3.\4 & \5.\6 & \7.\8 \\\\

\begin{table}
    \caption{Performances of the models with mean $\pm$ standard deviation, averaged on the population.}
    \label{table:res}
    \begin{tabularx}{\linewidth}{C||C|C|C}
        \toprule
        \textbf{Model} & \textbf{RMSE} & \textbf{MAPE} & \textbf{TL} \\
    
         \midrule
         
         \multirow{1}{*}{\textbf{RF}} & 19.23 $\pm$ 6.73 & 9.37 $\pm$ 1.58 & 15.31 $\pm$ 3.38\\
         
         \multirow{1}{*}{\textbf{LSTM}} & \textbf{17.52 $\pm$ 5.52} & \textbf{8.35 $\pm$ 1.30} & \textbf{12.01 $\pm$ 2.36}\\
         
         \multirow{1}{*}{\textbf{RETAIN}} & 17.60 $\pm$ 4.90 & 8.58 $\pm$ 0.84 & 12.14 $\pm$ 2.53\\

        \bottomrule
    \end{tabularx}
\end{table}

%% file: tables/results_cg_ega.tex
% (.*)\t(.*)\t0,(\d\d)(.*)\t0,(\d\d)(.*)\t0,(\d\d)(.*)
% \1 & \2 & \3.\4 & \5.\6 & \7.\8 \\\\

\begin{table}
    \caption{Clinical acceptability of the models with mean $\pm$ standard deviation, averaged on the population.}
    \label{table:res_cg_ega}
    
    \begin{tabularx}{\linewidth}{C||C|C|C}
        \toprule
        \multirow{2}{*}{\textbf{Model}} &  \multicolumn{3}{c}{\textbf{CG-EGA}} \\
        &  AP & BE & EP \\
         
         \midrule
         
         \multirow{1}{*}{\textbf{RF}} & \textbf{86.00 $\pm$ 4.37} & \textbf{10.79 $\pm$ 3.59} & 3.21 $\pm$ 0.84\\
         
         \multirow{1}{*}{\textbf{LSTM}} & 85.67 $\pm$ 3.28 & 11.46 $\pm$ 2.47 & \textbf{2.87 $\pm$ 0.95}\\
         
         \multirow{1}{*}{\textbf{RETAIN}} & 85.54 $\pm$ 5.41 & 11.56 $\pm$ 4.50 & 2.90 $\pm$ 0.95\\
         
        \bottomrule
    \end{tabularx}
\end{table}

%% file: 5-Conclusion.tex
\section{Conclusion}

In this study, we have presented the application of the RETAIN model proposed by Choi \textit{et al.} \cite{choi2016retain} to the challenging task of 30-minutes ahead-of-time glucose prediction for diabetic people. Using a two-level attention mechanism, the RETAIN model is able to produce interpretable predictions, which is highly valuable in the context of a biomedical field.

We have evaluated the model on a type-2 diabetic population of 5 patients and compared it against a Random Forest and a LSTM-based recurrent neural network. By being interpetable while respectively equalling and outperforming the LSTM and the RF models, we show that the RETAIN model is very promising. 

In the future, we plan to extend the study to another dataset, namely the Ohio T1DM dataset \cite{marling2018ohiot1dm}. In particular, this dataset comprises 6 type-1 diabetic patients with similar data. Also, thanks to the interpretability of the RETAIN model, we plan to explore variants of its architecture and input data (e.g., physical activity measures).

\section*{Acknowledgment}

We would like to thank the diabetes health network Revesdiab and Dr. Sylvie JOANNIDIS for their help in building the IDIAB dataset used in this study.

\bibliographystyle{splncs04}
\bibliography{refs.bib}